%% file: SimplexCoding_arxiv.tex
\documentclass[10pt]{article}

\usepackage[psamsfonts]{amssymb}
\usepackage{amsmath, amsthm, graphicx, enumerate, subfigure, color}
\usepackage{palatino}
\usepackage{color, palatino}
\usepackage[top=3cm, bottom=3cm, left=2cm, right=2cm]{geometry}
\usepackage{algorithmic}
\usepackage{algorithm}
\usepackage{tikz}
\input{macro}

\begin{document}
\title{Multiclass Learning  with Simplex Coding}

\author{Youssef Mroueh$^{\sharp,\ddagger}$, Tomaso Poggio$^{\sharp}$,   Lorenzo Rosasco$^{\sharp,\ddagger}$ Jean-Jacques E. Slotine${\dagger}$ \\
\small \em $\sharp$ - CBCL, McGovern Institute, MIT;\small \em $\dagger$ - IIT; \small \em $\dagger$ - ME, BCS, MIT\\
\small \tt ymroueh, lrosasco,jjs@mit.edu tp@ai.mit.edu}

\date \today


\maketitle
\begin{center}
\end{center}

\begin{abstract}
In this paper we discuss a novel  framework for multiclass learning, 
defined by  a suitable coding/decoding strategy,  namely the simplex coding, that allows to generalize to multiple classes
a relaxation approach commonly used in binary classification.
In this framework, a relaxation error analysis can be developed avoiding constraints on the considered hypotheses class. 
Moreover, we show that in this setting it is possible to derive the first provably consistent 
regularized method with training/tuning complexity which is {\em independent} to the number of classes.
Tools from convex analysis are introduced that can be used beyond the scope of this paper. 

\end{abstract}

\section{Introduction}

As bigger and more complex datasets are available, multiclass learning is becoming increasingly important in machine learning.
While theory and algorithms for solving binary classification problems are well established, the problem of multicategory classification is much less understood. Practical multiclass algorithms often reduce the problem to a collection of binary classification problems. Binary classification algorithms are often based 
on  a {\em relaxation approach}: classification is posed as a non-convex minimization problem and hence relaxed to a convex one, defined by suitable convex loss functions. In this context,   results in statistical learning theory  quantify  the error incurred by  relaxation and in particular derive {\em comparison inequalities} explicitly relating  the excess misclassification risk with the excess expected loss, see for example \cite{bajomc06,yaroca07,RW10,zhangT} and \cite{stch08} Chapter 3 for an exhaustive presentation as well as generalizations.\\
Generalizing the above approach and results to more than two classes is not straightforward.
Over the years,  several computational solutions have been proposed (among others, see  \cite{wahba,dietterich95solving,Singer,Weston, ASS00,tsochantaridis2005largemargin}.  Indeed, most of the above methods can be interpreted as a kind of relaxation. Most proposed
methods have complexity which is more than linear in the number of classes and  simple one-vs all in practice offers 
a good alternative both in terms of performance and speed  \cite{Rif}.
Much fewer works have focused on deriving theoretical guarantees. Results in this sense have been pioneered by \cite{zhang04stat,tewari05consistency}, see also \cite{fisher,g07,VWR11}. 
In these works the error due to relaxation is studied asymptotically and under  constraints on the function class to be considered.
More quantitative results in terms of comparison inequalities are given in \cite{chinese}  under similar restrictions 
(see also \cite{vandeGeer}). 
 Notably, the above  results  show that seemigly intuitive extensions of binary classification 
algorithms might lead to methods which are not consistent. 
Further, it is interesting to note that  these restrictions on the function class, needed to prove the theoretical guarantees,  make the computations in the corresponding algorithms more involved and are in fact  often  ignored in practice.\\
In this paper we dicuss a novel  framework for multiclass learning, 
defined by  a suitable coding/decoding strategy,  namely the simplex coding, 
in which a relaxation error analysis can be developed avoiding constraints on the considered hypotheses class. 
Moreover, we show that in this framework it is possible to derive the first  provably consistent 
regularized method with training/tuning complexity which is {\em independent} to the number of classes.
Interestingly, using the simplex coding, we can naturally generalize results, proof techniques and methods from the binary case, which is recovered as a special case of our theory. Due to space restriction in this paper we focus on extensions of least squares, and SVM loss functions, but our analysis can be generalized to large class of simplex loss functions, including extension of logistic and exponential loss functions (used in boosting). Tools from convex analysis are developed in the longer version of the paper  and can be useful beyond the scopes of this paper, and in particular in structured prediction.  
%
 
 The rest of the paper is organized as follow.  In Section \ref{back} we discuss problem statement and background.
 In Section \ref{sec:simp} we discuss the simplex coding framework that we analyze in  Section \ref{sec:theory}.
 Algorithmic aspects and numerical experiments are discussed in Section  \ref{sec:algo} 
 and  Section \ref{sec:exp}, respectively. Proofs and supplementary technical results are given in the longer version of the paper.

\section{Problem Statement and Previous Work}\label{back}
%

Let $(X,Y)$ be two random variables with values in two measurable spaces $\mathcal{X}$ and $\mathcal{Y}=\{1 \dots T \}$, $T\geq 2$. Denote by $\rho_{\XX}$, the law of $X$ on $\XX$, and by $\rho_j(x)$, the conditional probabilities for $j\in \mathcal{Y}$. The data is a sample $S=(x_i,y_i)_{i=1}^n$, from $n$ identical and independent  copies of $(X,Y)$.We can think of $\XX$ as a set of possible inputs and of $\mathcal{Y}$, as a set of labels describing a set of semantic categories/classes the input can belong to. A classification rule is a map $b: \XX \to \mathcal {Y}$, and its  error is measured by the misclassification risk
$R(b)=\mathbb{P}(b(X) \neq Y)=\mathbb{E}(\ind_{[b(x)\neq y]}(X,Y)).$
The optimal  classification rule that minimizes $R$ is the Bayes rule, $b_{\rho}(x)=\argmax_{y \in \YY} \rho_y(x), x \in \XX.$
Computing the Bayes rule by directly  minimizing the risk $R$, is not possible since the probability distribution is unknown.
In fact one could think of minimizing the empirical risk  (ERM), $R_S(b)=\frac 1 n \sum_{i=1}^n\ind_{[b(x)\neq y]}(x_i,y_i)$,
%
which is an unbiased estimator of the $R$, but  the corresponding 
optimization problem is in general  not feasible.
In binary classification, one of the most common way to obtain computationally efficient methods
is based on a relaxation approach. We recall this approach in the next section and describe its extension to multiclass in the rest of the paper.\\
\noindent {\bf Relaxation Approach to Binary Classification.}  If  $T=2$,  we can set  $\YY=\pm 1$.
Most modern machine learning algorithms for binary classification consider a convex relaxation of  the ERM functional $R_S$. More precisely:
1)  the indicator function in $R_S$ is replaced by non negative loss $V: \mathcal{Y}\times \mathbb{R}\to \mathbb{R^{+}}$ which is convex in the second argument and is  sometimes called a {\em surrogate} loss; 
2)  the classification rule $b$ replaced by a real valued measurable function $f: \XX \to \mathbb{R}$.
A classification rule is then obtained by considering the sign of $f$.
It often suffices to consider a special class of loss functions, namely 
large margin loss functions $V: \mathbb{R}\to \mathbb{R}^{+}$ of the form $V(-yf(x))$.
This last expression is  suggested by the observation that the misclassification risk, using the labels $\pm1$, can be written as $R(f)=\mathbb{E}(\Theta(-Yf(X))),$
where $\Theta$ is the heavy side step function. The quantity  $m=-yf(x)$, sometimes called  the {\em margin}, is a natural point-wise measure of the classification error.  Among other examples of large margin loss functions (such as the logistic and exponential loss), 
we recall the  hinge loss $V(m)=\hi{1+m}=\max\{1+m,0\}$ used in support vector machine, and  the square loss $V(m)=(1+m)^2$ used in regularized least squares (note that $(1-yf(x))^2=(y-f(x))^2$). 
Using surrogate large margin loss functions it is possible to  design effective learning algorithms replacing the empirical risk with  regularized empirical risk minimization
\begin{equation}\label{ERMV}
\EE^\la_S(f)=\frac{1}{n}\sum_{i=1}^nV(y_i,f(x_i))+\la {\cal R}(f),
\end{equation}
where $\cal R$ is a suitable regularization functional and $\la$ is the regularization parameter, see Section \ref{sec:algo}.

\subsection{Relaxation Error Analysis}{\label{sec:relax}}

As we replace the misclassification loss with a convex {\em surrogate|} loss, we are effectively changing the problem: 
 the misclassification risk is replaced by the expected loss,  
$\mathcal{E}(f)=\mathbb{E} (V(-Yf(X)))$
. The expected loss  can be seen as a functional on a large  space of functions 
${\cal F}={\cal F}_{V,\rho}$, which depend on $V$ and $\rho$. Its minimizer, denoted by $f_\rho$, replaces the Bayes rule as the target of our algorithm.\\
The  question arises  of the price we pay by a considering a relaxation approach: 
``What is the relationship between $f_\rho$ and $b_\rho$?'' More generally, ``What is the approximation we incur into
by  estimating  the expected risk rather   than  the misclassification risk?''
The {\em relaxation error} for a given loss function can be  quantified by the following  two requirements:\\
\noindent {1) \emph{Fisher Consistency}}. A loss function is Fisher consistent if   $\text{sign}(f_{\rho}(x))=b_{\rho}(x)$ almost surely (this property is related to the notion of  classification-calibration  \cite{bajomc06}).\\
\noindent { 2)  \emph{Comparison inequalities}}.  The excess misclassification risk, and  the excess expected loss are related by a  comparison inequality 
$$R(\text{sign}(f))-R(b_{\rho}) \leq \psi(\mathcal{E}(f)-\mathcal{E}(f_\rho)),$$ 
for any function $f\in \cal F$, where $\psi=\psi_{V,\rho}$ is a suitable function  that depends on $V$, and possibly on the data distribution. 
In particular $\psi$ should be such that  $\psi(s)\to 0$ as $s\to0$,  so that   if $f_n$ is a (possibly random) sequence of functions, such that $\mathcal{E}(f_n)\to \mathcal{E}(f_{\rho})$ (possibly  in probability), then the corresponding sequences of classification rules $c_n=\text{sign}(f_n)$ is Bayes consistent, i.e. $R(c_n)\to R(b_{\rho})$ (possibly  in probability). If $\psi$ is explicitly known, then bounds on the  excess expected loss  yields bounds on the excess misclassification risk.\\
The relaxation error in the binary case has been thoroughly studied in \cite{bajomc06, RW10}. 
In particular,  Theorem 2 in  \cite{bajomc06} shows  that if a  large margin surrogate
loss is convex, differentiable and decreasing in a neighborhood of $0$, then the loss is Fisher consistent. 
Moreover, in this case it is possible to give an explicit expression of the function $\psi$.
In particular, for the hinge loss the target function is exactly the Bayes rule and $\psi(t)=|t|$. For least squares, $f_{\rho}(x)=2\rho_1(x)-1$, and $\psi(t)=\sqrt{t}$.
The comparison  inequality for the square loss can be improved for a suitable class of probability distribution satisfying  the so called Tsybakov noise condition \cite{T04}, $\rho_\XX(\{x \in \XX, |f_{\rho}(x)|\leq s\})\leq B_q s^q ,s \in [0,1], q> 0.$
 Under this condition the probability of points such that $\rho_y(x)\sim \frac{1}{2}$ decreases polynomially.
In this case the comparison inequality for the square loss is given by $\psi(t)=c_{q} t^{\frac{q+1}{q+2}}$, see  \cite{bajomc06,yaroca07}.\\
\noindent {\bf Previous Works in Multiclass Classification.}
From a practical perspective, over the years,  several computational solutions to multiclass learning have been proposed.
Among others, we mention  for example \cite{wahba,dietterich95solving,Singer,Weston, ASS00,tsochantaridis2005largemargin}. 
Indeed, most of the above methods can be interpreted as   a kind of relaxation of the original multiclass problem. 
Interestingly,  the study in  \cite{Rif}  suggests that the  simple one-vs all schemes should be a practical benchmark for multiclass algorithms
as it seems to experimentally achive performances that are similar or better to  more sophisticated methods.\\
As we previously mentioned from a theoretical perspective a general account of a large class of multiclass methods 
has been given in  \cite{tewari05consistency}, building on  results in \cite{bajomc06} and \cite{zhang04stat}.
Notably, these  results  show that seemingly intuitive extensions of binary classification algorithms might lead to {\em inconsistent} methods. 
These results, see also \cite{fisher,VWR11}, 
are developed in a setting where  a classification rule is found by applying a suitable prediction/decoding map  to a function $f:\XX\to \R^T$ 
where $f$ is found considering a loss function  $V:\YY \times \R^T\to \R^+.$
The considered  functions have to satisfy the constraint  $\sum_{y\in \YY} f^y(x)=0$, for all $x\in \XX$.
The latter requirement is problematic since it makes the computations in the corresponding algorithms more involved 
and is in fact  often  ignored, so that  practical algorithms often come with no consistency guarantees. 
In all  the above papers relaxation  is studied in terms of  Fisher and Bayes consistency
and the explicit form of the function $\psi$ is not given. More quantitative results in terms of explicit comparison inequality are given in \cite{chinese} and (see also \cite{vandeGeer}), but  also need to to impose the "sum to zero" constraint on the considered function class.

\section{A Relaxation Approach to  Multicategory Classification}\label{sec:simp}
In this section we propose a natural extension of the relaxation approach
that avoids  constraining the class of functions to be considered, and allows to derive explicit 
comparison inequalities. See Remark~\ref{presimplex} for related approaches.
\paragraph{Simplex Coding.}\label{theory}
 \begin{figure}[t]
\begin{center}
\begin{tikzpicture}[scale=0.35]
  
  \draw (-3.5,0) -- (3.5,0);
  \draw (0,-3.5) -- (0,3.5);
  \draw (0,0) circle (2 cm);
 \coordinate [label=right:\textcolor{blue}{$c_1$}]  (A) at (2,0);
 \coordinate [label=left:\textcolor{blue}{$c_2$}]  (B) at (-1,1.732);
 \coordinate [label=left:\textcolor{blue}{$c_3$}]  (C) at (-1,-1.732);
 \coordinate [label=right:\textcolor{blue}{$\alpha$}]  (M) at (1.414,1.414);
  \coordinate (E) at (-3,0);
 \coordinate  (F) at (1.5,-2.59);
 \coordinate  (G) at (1.5,2.59);
 \coordinate(O) at (0,0);
 \draw [->,very thick] (0,0) -- (A);
  \draw [->,very thick] (0,0) -- (B);
  \draw [->,very thick] (0,0) -- (C);
  \draw [->,very thick,red] (0,0) -- (M);
 \draw [ -,dashed,red] (A) -- (M);
 \draw [ -,dashed,red] (B) -- (M);
  \draw [-,dashed,red] (C) -- (M);
\end{tikzpicture}
\end{center}
\caption{Decoding with simplex coding $T=3$.}
\label{fig:simplex}
\end{figure}
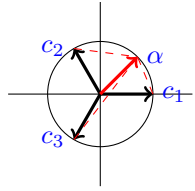
We start considering a suitable coding/decoding strategy.
A {\em coding} map  turns a   label $y\in \YY$ into a code vector.
The corresponding  {\em decoding} map  given a vector returns a label in $\cal Y$. 
Note that, this is what we implicitly did  while treating  binary classification {\em encoding} the label space
$\YY=\{1, 2\}$ using the coding $\pm 1$, so that the naturally decoding strategy is simply $\text{sign}(f(x))$.
%
The coding/decoding strategy  we study is described by the following definition.
 \begin{definition}[{\bf Simplex Coding}]\label{scode}
The simplex coding is a map $C:\YY \to \R^{T-1},$ $\quad C(y)=\a_y,$
where the code vectors ${\AA}=\{c_y~|~y\in {\cal Y}\}\subset  \R^{T-1}$ satisfy: 
1) $\norT{\a_y}^2=1$, $\forall  y\in \YY$, 2)$\scalT{\a_y}{\a_{y'}}=-\frac{1}{T-1},$ for $ y\neq y'$ with $y,y' \in \YY$, and 
3) $\sum_{y\in \YY} \a_y =0$.
The corresponding decoding is the map 
$D: \R^{T-1}\to \{1, \dots, T\}, \quad\quad  D(\alpha)=\argmax_{y\in \YY}\scalT{\alpha}{c_y},$ $\forall \alpha \in \R^{T-1}.$
\end{definition}
The simplex coding corresponds to the $T$ most separated  vectors on the 
hypersphere $\mathbb{S}^{T-2}$ in $\mathbb{R}^{T-1}$, that is the vertices of the simplex (see Figure \ref{fig:simplex}). 
For binary classification it reduces to the  $\pm 1$ coding and the decoding map is equivalent to taking the  sign of $f$.
The decoding map has a natural geometric interpretation: an input  point is mapped to a vector $f(x)$ by a function $f:\XX\to \R^{T-1}$,
  and  hence  assigned to the class having closer code 
 vector (for $y,y'\in \YY$ and $\alpha \in \R^{T-1}$, we have $\nor{\a_y-\alpha}^2\ge \nor{\a_{y'}-\alpha }^2\Leftrightarrow \scal{\a_{y'}}{\alpha}\le \scal{\a_y}{\alpha}$. \\
\noindent{\bf Relaxation for Multiclass Learning.}
We use the simplex coding to  propose an extension of the binary classification approach. 
 Following the binary case, the relaxation can be described in two steps:
 \begin{enumerate}
 \item using the simplex coding, the indicator function is  upper bounded  by a non-negative  loss function  $V: \mathcal{Y}\times \mathbb{R}^{T-1}\to \mathbb{R}^+$, such that $\ind_{[b(x)\neq y]}(x,y)\le V(y, C(b(x))), $for all   $b:\XX\to \YY$, and  $x\in \XX,y\in \YY$, 


\item rather than $C\circ b$ we consider functions with values in $f:\XX\to \R^{T-1}$, so that $V(y, C(b(x)))\le  V(y, f(x))$,
for all   $b:\XX\to \YY, f:\XX\to \R^{T-1}$ and  $x\in \XX,y\in \YY$.

\end{enumerate}

%
In the next section we discuss several loss functions satisfying the above definitions and we study in particular  the extension of the least squares and SVM loss functions.\\
\noindent{\bf Multiclass Simplex Loss Functions.}
 Several loss functions for binary classification can be naturally extended to multiple classes using the simplex coding. Due to space restriction, in this paper we focus on extensions of least squares, and SVM loss functions, but our analysis can be generalized to large class of simplex loss functions, including extension of logistic and exponential loss functions( used in boosting). The Simplex Least Square loss (\textbf{S-LS}) is given by $V(y,f(x))=\nor{c_y-f(x)}^2$,
 and reduces to the usual  least square approach to binary classification  for $T=2.$
One  natural extension of the SVM's hinge loss in this setting would be to consider the Simplex Half space SVM loss (\textbf{SH-SVM})
$V(y,f(x))=\hi{1-\scalT{\a_y}{f(x)}}$.
We will see in the following that while this loss function
 would induce   efficient algorithms in general is not Fisher consistent unless further constraints are assumed.
 In turn, this latter constraint would considerably slow down the computations. 
 Then we consider a second  loss function Simplex Cone SVM (\textbf{SC-SVM}), related to the hinge loss, which is defined as $V(y,f(x))=\sum_{y'\neq y}\hi{\frac{1}{T-1}+\scalT{\a_{y'}}{f(x)}}.$
The latter loss function is related to the one considered in the multiclass SVM proposed in \cite{wahba}.
We will see that it is possible to quantify the relaxation error of the loss function without requiring further constraints. 
Both the above SVM loss functions reduce  to the binary SVM hinge loss if $T=2$. 
 \begin{remark}[Related approaches]\label{presimplex}
The simplex coding has been  considered in \cite{hd07},\cite{wl10}, and \cite{boosting}.
In particular,  a kind of SVM loss is considered in  \cite{hd07}
where $V(y, f(x))=\sum_{y'\neq y}\hi{\eps-\scal{f(x)}{v_{y'}(y)}}$
and $v_{y'}(y)=\frac{c_y-c_{y'}}{\nor{c_y-c_{y'}}},$
with $\eps=\scal{c_y}{v_{y'}(y)}=\frac{1}{\sqrt{2}}\sqrt{\frac{T}{T-1}}$.
 More recently \cite{wl10} considered the loss function $
V(y, f(x))=\hi{\eps-\nor{c_y-f(x)}}$,
and a simplex multi-class boosting loss was introduced in \cite{boosting}, in our notation $V(y,f(x))=\sum_{j\neq y}e^{-\scalT{\a_{y}-\a_{y'}}{f(x)}}.$ While all those losses introduce a certain notion of margin that makes use of the geometry of the simplex coding, it is not to clear how to derive explicit comparison theorems and moreover the computational complexity  of the resulting algorithms scales linearly with the number of classes  in  the case of the losses considered in \cite{boosting, wl10} and $O((nT)^\gamma),\gamma \in\{2,3\}$ for losses considered in \cite{hd07} . 
 
  
\end{remark}

\begin{figure}[H]
\begin{center}
\includegraphics[height=0.35\textwidth]{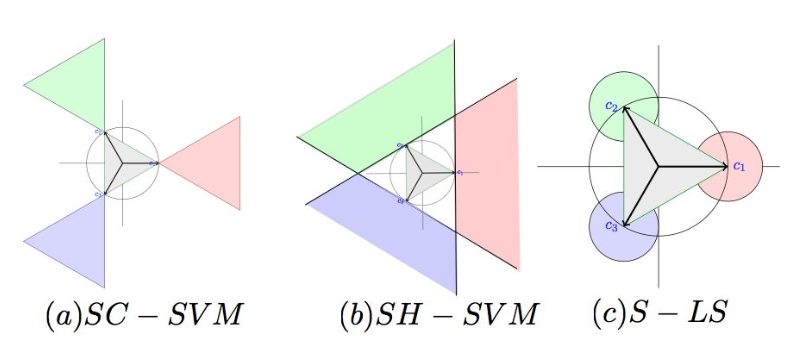}
\end{center}
\caption{Level sets of different losses considered for $T= 3$. A classification is correct if an input $(x,y)$ is mapped to a point $f(x)$ that lies in the neighborhood of the vertex $c_y$. The shape of the neighborhood is defined by the loss, it takes form of a cone supported on a vertex in the case of SC-SVM, a half space delimited by the hyperplane orthogonal to the vertex in the case of the SH-SVM,  and a sphere centered on the vertex in the case of S-LS.}
\end{figure}

\section{Relaxation Error  Analysis}\label{sec:theory}
If we consider the simplex coding, a function $f$ taking values in $\R^{T-1}$, and the decoding operator $D$,
the misclassification risk can also be written as:
$R(D(f))=\int_{\XX}(1-\rho_{D(f(x))})d\rho_{\XX}(x)$.
Then, following a relaxation approach we  replace the misclassification loss   by 
the expected risk induced by one of the  loss functions $V$ defined in the previous section.
 As in the binary case we consider the expected loss 
$
\mathcal{E}(f)=\int V(y,f(x))d\rho(x,y).
$
Let $L^p(\XX, \rho_\XX)=\{f:\XX \to \mathbb{R}^{T-1}~|~ \nor{f}_\rho^p=\int \nor{f(x)}^p d\rho_\XX(x)<\infty \}$, $p\geq1.$

The following theorem studies  the relaxation error for SH-SVM, SC-SVM, and S-LS loss functions.
\begin{theorem}
For SH-SVM, SC-SVM, and S-LS loss functions, there exists a $p$ such that  $\EE: \LLp \to \R^+$ is  convex and continuous. Moreover, 
\begin{enumerate}
\item The minimizer $f_\rho$ of $\EE$ over ${\cal F}= \{f \in \LLp~|~ f(x) \in K ~a.s.\}$ exists and 
$D(f_\rho)=b_\rho$.
\item For any $f\in {\cal F}$,
$
R(D(f))-R(D(f_{\rho}))\leq C_T( \mathcal{E}(f)-\mathcal{E}(f_{\rho}))^{\alpha},
$
where the expressions of $p,K,f_{\rho},C_T,$ and $\alpha$ are given in Table~\ref{table}.
\end{enumerate}
\begin{table}[H]
\begin{center}
    \begin{tabular}{|l|l|l|l|l|l| }
    \hline
Loss &$p$&$K$ &$ f_{\rho}$& $C_T$ & $\alpha$ \\ \hline
 SH-SVM &$1$&$conv({\cal C})$&$\a_{b_{\rho}}$&$T-1$&$1$\\ \hline
 SC-SVM &$1$& $\mathbb{R}^{T-1}$ &$\a_{b_{\rho}}$& $T-1$& $1$\\     \hline
 S-LS& $2$ &$\mathbb{R}^{T-1}$&$\sum_{y\in \YY} \rho_y \a_y$&$\sqrt{\frac{2(T-1)}{T}}$ & $\frac{1}{2}$\\ \hline
 \end{tabular}
  \label{table}
 \caption{ $conv(\mathcal{C})$ is the convex hull of the set $\mathcal{C}$ defined in \eqref{scode}.}

\end{center}
  \end{table}
\label{theo:summary}
\end{theorem}
\vspace{-0.5cm}
The proof of this theorem is given in the longer version of the paper.\\
The above theorem  can be improved for Least Squares under certain classes of distribution .
Toward this end we introduce the following notion of misclassification noise that generalizes Tsybakov's noise condition. 
\begin{definition}
Fix $q>0$, we say that the distribution $\rho$ satisfy the multiclass noise condition with parameter $B_q$, if
\begin{equation}\label{GenTsy1}
\rho_{\mathcal{X}}\left(\left\{x\in \mathcal{X}~|~  0 \leq \min_{j\neq D(f_{\rho}(x))}\frac{T-1}{T} ( \scalT{\a_{D(f_{\rho}(x))}-\a_j}{f_{\rho}(x)})\leq s\right \}\right)\leq B_q s^q,
\end{equation}
where  $s\in[0,1]$.
\end{definition}

If a distribution $\rho$ is characterized by  a very large $q$, then,  for each $x\in \XX$,  $f_\rho(x)$ 
is arbitrarily close to one of the coding vectors.
For $T=2$, the above condition reduces to the binary Tsybakov noise. Indeed, let $\a_1=1$, and $\a_2=-1$, 
if $f_{\rho}(x)>0$, $ \frac{1}{2} (\a_1-\a_2) f_{\rho}(x)= f_{\rho}(x)$, and  if $f_{\rho}(x)<0$, $\frac{1}{2}(\a_2-\a_1)f_{\rho}(x)=-f_{\rho}(x)$.

The following result improves the exponent of simplex-least square to $\frac{q+1}{q+2}>\frac{1}{2}$ :  

\begin{theorem}
For each $f\in L^2(\XX,\rho_{\mathcal{X}})$, if $(\ref{GenTsy1})$ holds, then for S-LS we have  the following inequality,
\begin{equation}
R(D(f))-R(D(f_{\rho}))\leq K \left(\frac{2(T-1)}{T}(\mathcal{E}(f)-\mathcal{E}(f_{\rho}))\right)^{\frac{q+1}{q+2}},
\label{eq:betterrate}
\end{equation}
for a constant $K =\left(2 \sqrt{B_q+1}\right)^{\frac{2q+2}{q+2}}$.
\label{pro:tsyb1}
\end{theorem}
\begin{remark}
Note that the comparison inequalities show a tradeoff between the exponent $\alpha$ and the constant $C(T)$, for S-LS and SVM losses.
While the constant is order $T$ for SVM it is order $1$ for S-LS, on the other hand the exponent is $1$ for SVM losses and $\frac{1}{2}$ for S-LS. The latter could be enhanced to $1$ for close to separable classification problems by virtue of the Tsybakov noise condition.   
\end{remark}
\begin{remark}
Comparison inequalities given in Theorems \ref{theo:summary} and \ref{pro:tsyb1} can be used to derive generalization bounds on the excess misclassification risk. For least square min-max sharp bound, for vector valued regression are easy to derive.
Standard techniques  for deriving sample complexity bound in binary classification extended for multi-class SVM losses  could be found in \cite{g07} and could be adapted to our setting. The obtained bound are not known to be tight, better bounds akin to those in \cite{stch08}, will be subject to future work. 
\end{remark}

\section{Computational Aspects and Regularization Algorithms}\label{sec:algo}

In this section we discuss some computational implications of the framework we presented.

 \noindent{\textbf{Regularized Kernel Methods.}} 
We  consider regularized methods of the form~\eqref{ERMV}, induced by  simplex loss functions and where 
the hypotheses space is  a vector valued  reproducing kernel Hilbert spaces (VV-RKHSs) and the regularizer the corresponding norm.
See Appendix D.$2$ for a brief introduction to VV-RKHSs. \\
In the following, we consider a class of kernels such that the  corresponding RKHS $\hh$
is given by the completion of the span  $\{f(x)=\sum_{i=j}^N \Gamma(x_j,x)\c_j,~\c_j\in \R^{T-1}, x_i, \in \XX,~\forall j=1, \dots, N\}$, where we note that the coefficients are vectors in $\R^{T-1}$. While other choices are possible this is the kernel more directly related to a one vs all approach. 
We will discuss in particular the case where the kernel is induced by a finite dimensional feature map, 
$k(x,x')=\scal{\Phi(x)}{\Phi(x')}, \quad  \text{where}\quad  \Phi:\XX\to \R^p$, and $\scal{\cdot}{\cdot}$ is the inner product in $\R^p$.
In this case we can write each function in $\hh$ as $f(x)=W\Phi(x)$, where  $W\in \R^{(T-1)\times p}$. \\
It is known \cite{micpon05,capdev05} that the representer theorem \cite{wahba70} can be easily extended to a vector valued setting, so that 
that  minimizer of a simplex version of  Tikhonov regularization is given by  $f_S^\la(x)=\sum_{j=1}^n k(x,x_j)\c_j$, $ \c_j \in \mathbb{R}^{T-1},$ for all $x\in \XX$, where the explicit expression of the coefficients depends on the considered loss function. 
We use the following notations:
$K\in \mathbb{R}^{n \times n},K_{ij}= k(x_i,x_j), \forall i,j \in \{1 \dots n\}, A \in \mathbb{R}^{n\times (T-1)} , \A=(\c_1,...,\c_n)^T.$\\
\noindent{\bf Simplex Regularized Least squares (S-RLS).}
S-RLS is obtained considering the simplex least square loss in the Tikhonov functionals.
It is easy to see  \cite{Rif} that in this case the coefficients must satisfy
either $(K+\lambda n I)A=\Y$ or $(\Xn^T\Xn+\la n I) W =\Xn^T\Y$ in the linear case, where $\Xn \in \mathbb{R}^{n\times p} , \Xn=(\Phi(x_1),...,\Phi(x_n))^{\top}$ and $\Y \in \mathbb{R}^{n\times (T-1)}, \Y=(\a_{y_1},...,\a_{y_n})^{\top}$ .\\
Interestingly, the classical results from \cite{wahba90} can be extended to show that 
the value $f_{S_i}(x_i)$, obtained computing  the solution $f_{S_i}$  removing the $i-th$ point 
from the training set (the leave one out solution),  can be computed in closed form.
Let $f_{loo}^{\lambda}\in \mathbb{R}^{n\times (T-1)}, f_{loo}^{\lambda}=(f_{S_1}^{\lambda}(x_1),\dots,f_{S_n}^{\lambda}(x_n))$.
Let $\mathcal{K}(\lambda)=(K+\lambda n I)^{-1}$and $C(\lambda)=\mathcal{K}(\lambda)\hat{Y}$.
Define $M(\lambda)\in \mathbb{R}^{n \times (T-1)}$, such that: $M(\lambda)_{ij}=1/\mathcal{K}(\lambda)_{ii}$, $\forall~j=1\dots T-1.$
One can show similarly to  \cite{Rif}, that $f_{loo}^{\lambda}=\hat{Y}-C(\lambda)\odot M(\lambda)$, where $\odot$ is the Hadamard product.
Then,  the leave-one-out  error 
$\frac 1 n  \sum_{i=1}^n \ind_{y\neq D(f_{S^i}(x)) }(y_i, x_i),$
can be minimized  at essentially no extra cost by precomputing the eigen decomposition of  $K$ (or $\Xn^T\Xn$).\\
\noindent{\bf Simplex Cone Support Vector Machine (SC-SVM).} Using standard reasoning  it is easy to show that (see Appendix C.$2$), 
for the SC-SVM the coefficients in the representer theorem are given by  
$\c_i=-\sum_{y\neq y_i}\alpha^{y}_i \a_y, \quad i=1, \dots, n,$
where  $\alpha_i=(\alpha^y_i)_{y\in \YY} \in \R^T, i=1, \dots, n,$  solve the quadratic programming (QP) problem 
\begin{eqnarray}\label{SCSVMQP}
&&\max_{\alpha_1, \dots, \alpha_n \in \R^T } \left\{ -\frac{1}{2}\sum_{y,y',i,j}\alpha^y_i  K_{ij}  G_{yy'} \alpha ^{y'}_j+\frac{1}{T-1} \sum_{i=1}^n \sum_{y=1}^T\alpha^y_i\right\}\\
&& \text{subject to}\quad 0\leq \alpha_i^y \leq C_0\delta_{y, y_i}, ~\forall~i=1, \dots, n, y\in \YY\nonumber
\end{eqnarray}
where $G_{y,y'}=\scalT{\a_y}{\a_{y'}}  \forall  y ,y' \in \mathcal{Y}$ and $C_0=\frac{1}{2n\la}$,  $\alpha_i=(\alpha^y_i)_{y\in \YY} \in \R^T$, for $i=1, \dots, n$ and $\delta_{i,j}$ is the Kronecker delta.\\
\noindent{\bf Simplex Halfspaces Support Vector Machine (SH-SVM).}  A similar, yet more  more complicated procedure, can be derived for the SH-SVM.
Here,  we omit this derivation and  observe instead that  if we neglect the convex hull 
constraint from Theorem~\ref{theo:summary}, requiring $f(x)\in \text{co}(\AA)$ for almost all $x\in \XX$,  
then the SH-SVM  has an especially simple formulation at the price of  loosing consistency guarantees.
In fact, in this case the coefficients  are given by  
$\c_i= \alpha_i \a_{y_i}, \quad i=1, \dots, n,$
where $\alpha_i\in \mathbb{R}$, with  $ i=1, \dots, n$   solve the quadratic programming (QP) problem 
\begin{eqnarray*}
&&\max_{\alpha_1, \dots, \alpha_n \in \R} -\frac{1}{2}\sum_{i,j} \alpha_i K_{ij}G_{y_iy_j}\alpha_j +\sum_{i=1}^n \alpha_i\\
&& \text{subject to}\quad0\leq \alpha_i \leq C_0, ~\forall~ i =1 \dots n,
\end{eqnarray*}
where  $C_0=\frac{1}{2n\la}$. The latter formulation could be trained at the same complexity of the binary SVM (worst case $O(n^3)$) but lacks consistency.\\
\noindent {\bf Online/Incremental Optimization}
The regularized estimators induced by the simplex loss functions can be computed 
by mean of online/incremental first order (sub) gradient methods.  
Indeed, when considering finite dimensional feature maps, 
these strategies offer computationally feasible solutions to train estimators 
for  large datasets where neither  a $p$ by $p$ or an $n$ by $n$ matrix fit in memory.
Following \cite{pegasos} we can alternate  a step of  stochastic descent on a data point : $W_{\text{tmp}}= (1-\eta_i \lambda)W_i-\eta_i \partial(V(y_i, f_{W_i}(x_i)))$ and a projection on the Frobenius ball $W_i=\min(1, \frac{1}{\sqrt{\lambda}||W_{tmp}||_{F}})W_{\text{tmp}}$
(See Algorithn C.$5$ for details.)
The algorithm depends on the used loss function through the computation 
of the (point-wise) subgradient $\partial(V)$. The latter can be easily computed for all the loss functions previously discussed. 
For the SLS loss we have 
$ \partial(V(y_i, f_{W}(x_i)))=2  (\a_{y_i}-Wx_{i}) x_{i}^{\top},$
while for the SC-SVM loss we have 
$
 \partial(V(y_i, f_{W}(x_i)))= (\sum_{k\in I_i} c_k)x_i^{\top}
$
where  $I_i=\{y\neq y_i | \scalT{c_y}{Wx_i}>-\frac{1}{T-1}\}$.
For the SH-SVM loss we have: 
$
 \partial(V(y, f_{W}(x_i)))= - c_{y_i}x_i^{\top} \text{ if } c_{y_i}Wx_i<1 \text{ and } 0 \text{  else }.
$
\subsection{Comparison of Computational Complexity}
The cost  of solving S-RLS for fixed $\lambda$ is in the worst case $O(n^3)$ (for example via Choleski decomposition). 
If we are interested into computing the regularization  path for $N$ regularization parameter values, then as noted in \cite{Rif} it   might be convenient to perform an eigendecomposition of the kernel matrix rather than solving the systems $N$ times.
For explicit feature maps the cost is $O(np^2)$, so that  the cost of computing  the regularization path for  simplex RLS algorithm is $O(min(n^3,np^2))$ and hence  {\em independent} of $T$. One can contrast this complexity with the one of a n\"aive One Versus all  (OVa) approach that would lead to a 
$O(Nn^3T)$ complexity. Simplex SVMs can be  solved using  solvers available for binary SVMs that are considered to  
have complexity $O(n^\gamma)$ with  $\gamma \in \{2,3\}$(actually the complexity scales with the number of support vectors) . For SC-SVM, though, we have $nT$ rather than $n$ 
unknowns and the complexity is $(O(nT)^{\gamma})$.  
SH-SVM where we omit the constraint, could be trained at the same complexity of the binary SVM (worst case $O(n^3)$) but lacks consistency. Note that unlike for  S-RLS, there is no straightforward way to 
compute the regularization path and the leave one out error for any of the above SVMs . The online algorithms induced by the different simplex loss functions are essentially 
the same, in particular each iteration depends linearly on the number of classes.
\section{Numerical Results}{\label{sec:exp}}
We conduct several experiments to evaluate the performance of our batch and online algorithms, on 5 UCI datasets as listed in Table \ref{datas}, as well as  on  Caltech101 and Pubfig83. 
We compare the performance of our algorithms to on versus all svm (libsvm) , as well as the simplex based boosting \cite{boosting}.
For UCI datasets we use the raw features, on Caltech101 we use hierarchical features\footnote{The data set will be made available upon acceptance.} , and on Pubfig83 we use the feature maps from \cite{PintoEtAl2011}. 
In all cases the parameter selection is based either on a hold out (ho) $(80 \% \text{ training }- 20\% \text{ validation})$ or a leave one out error (loo).
For the  model Selection of $\lambda$ in S-LS,  $100$ values are chosen in the range $[\lambda_{min},\lambda_{max}]$,(where $\lambda_{min}$ and $\lambda_{\max}$, correspond to the smallest and biggest eigenvalues of $K $).
In the case of a Gaussian kernel (rbf) we use a heuristic that sets the width of the gaussian $\sigma $ to the 25-th percentile of pairwise distances between distinct points in the training set.  
In Table \ref{datas} we collect the resulting classification accuracies:

\begin{table}[H]\label{datas}
\begin{center}
  \small
    \begin{tabular}{|l|l|l|l|l|l|l|l| }

    \hline&Landsat & Optdigit &Pendigit& Letter &Isolet&Ctech&Pubfig83 \\ \hline
SC-SVM \tiny{ Online (ho)}&$65.15  \%$&$89.57 \%$  &$81.62\%$& $52.82\%$& $88.58\%$&$63.33\%$&$84.70\%$\\     \hline
SH-SVM \tiny{Online (ho)} &$75.43 \%$&$85.58\%$&$72.54\%$&$38.40\%$&$77.65\%$&$45\%$&$49.76\%$\\ \hline
S-LS \tiny{ Online (ho)}& $63.62 \%$ &$91.68\%$&$81.39\%$&$54.29\%$ &$92.62\%$ &$58.39\%$&$83.61\%$\\ \hline
S-LS   \tiny{Batch (loo)}& $65.88 \%$ &$91.90\%$&$80.69\%$&$54.96\%$ & $92.55\%$&$66.35\%$&$86.63\%$\\ \hline
S-LS rbf \tiny{ Batch (loo)}& $\bf{90.15 \%}$ &$\bf{97.09\%}$&$\bf{98.17}\%$&$\bf{96.48\%}$ &$\bf{97.05\%}$&$\bf{69.38\%}$&$\bf{86.75\%}$\\ \hline
SVM  \tiny{ batch ova (ho)}& $72.81 \%$ &$92.13\%$&$86.93\%$&$62.78\%$ &$90.59\%$&$70.13\%$&$85.97\%$\\ \hline
SVM  rbf\tiny{ batch ova (ho)}& $95.33 \%$ &$98.07\%$&$98.88\%$&$97.12\%$&$96.99\%$ &$51.77\%$&$85.60\%$\\ \hline
Simplex boosting \cite{boosting}& $ 86.65\%$ &$92.82\%$&$92.94\%$&$59.65\%$ &$91.02\%$&$-$&$-$\\ \hline
    \end{tabular}
    \end{center}
    \caption{Accuracies of our algorithms on several datasets.}
    \end{table}
    \vspace{-0.5 cm}
As suggested by the theory, the consistent methods SC-SVM and S-LS have a big advantage over SH-SVM (where we omitted the convex hull constraint) . Batch methods are overall superior to online methods, with online SC-SVM achieving the best results.
More generally, we see that rbf S- LS has the best performance among the simplex methods including the simplex boosting \cite{boosting}. When compared to One Versus All SVM-rbf, we see that S-LS rbf achieves essentially the same performance. 
%
%
%
%
%
%
%
%
%
%

\bibliographystyle{plain}
\bibliography{simplex}
\end{document}

%% file: macro.tex
\makeatletter
\def\@makechapterhead#1{
\vspace*{80\p@}
{\parindent \z@ \raggedright \normalfont
\ifnum \c@secnumdepth >\m@ne
\if@mainmatter
\huge\bfseries \@chapapp\space \thechapter
\par\nobreak
\vskip 20\p@
\fi
\fi
\interlinepenalty\@M
\Huge \bfseries #1\par\nobreak
\vskip 40\p@
}}
\makeatother
\newcommand{\ind}{1{\hskip -2.5 pt}\hbox{I}}

\def\argmax{\operatornamewithlimits{arg\,max}}

\renewcommand{\a}{c}
\renewcommand{\c}{a}
\renewcommand{\AA}{{\cal C}}

\newcommand{\iii}{\begin{enumerate}}
\newcommand{\fff}{\end{enumerate}}
\newcommand{\iiii}{\begin{itemize}}
\newcommand{\ffff}{\end{itemize}}
\newcommand{\mfi}{\begin{eqnarray*}}
\newcommand{\mff}{\end{eqnarray*}}
\newcommand{\mfni}{\begin{eqnarray}}
\newcommand{\mfnf}{\end{eqnarray}}

\newtheorem{theorem}{Theorem}
\newtheorem{definition}{Definition}

\newtheorem{remark}{Remark}

\providecommand{\nor}[1]{\left\lVert {#1} \right\rVert}

\providecommand{\scal}[2]{\left\langle{#1},{#2}\right\rangle}

\providecommand{\scalT}[2]{\left\langle{#1},{#2}\right\rangle}
\providecommand{\norT}[1]{\left\lVert {#1} \right\rVert}

\providecommand{\hi}[1]{\left|{#1} \right|_+}

\newcommand{\R}{\mathbb R}

\newcommand{\XX}{\mathcal X}
\newcommand{\hh}{\mathcal H}
\newcommand{\EE}{\mathcal E}
\newcommand{\A}{A}
\newcommand{\la}{\lambda}
\newcommand{\YY}{\mathcal Y}

\newcommand{\LLp}{L^p(\XX, \rho_\XX)}

\newcommand{\Y}{\hat{Y}}
\newcommand{\Xn}{\hat{X}}

\newcommand{\eps}{\varepsilon}